\documentclass[10pt,twocolumn,twoside]{IEEEtran}

\ifCLASSOPTIONcompsoc
  \usepackage[nocompress]{cite}
\else
  \usepackage{cite}
\fi

\ifCLASSINFOpdf
  \usepackage[pdftex]{graphicx}
\else
  \usepackage[dvips]{graphicx}
\fi

\usepackage{amsthm}
\usepackage{amsmath}
\usepackage{booktabs}
\usepackage{bm}
\usepackage{multirow}
\usepackage{enumitem}
\usepackage{amssymb}
\usepackage{makecell}
\usepackage{xcolor}
\usepackage{url}
\allowdisplaybreaks
\ifCLASSOPTIONcompsoc
  \usepackage[caption=false,font=normalsize,labelfont=sf,textfont=sf]{subfig}
\else
  \usepackage[caption=false,font=footnotesize]{subfig}
\fi

\newtheorem{assumption}{Assumption}

\begin{document}
\title{Loss re-scaling VQA: Revisiting the Language Prior Problem from a Class-imbalance View}

\author{Yangyang Guo,
        Liqiang Nie*~\IEEEmembership{Senior Member,~IEEE},
        Zhiyong Cheng,
        Qi Tian~\IEEEmembership{Fellow,~IEEE},
        Min Zhang~\IEEEmembership{Member,~IEEE}
\IEEEcompsocitemizethanks{
\IEEEcompsocthanksitem This work is supported by the National Natural Science Foundation of China, No.:U1936203. The associate editor coordinating the review of this manuscript and approving it for publication was Prof. Hengtao Shen. (Corresponding author: Liqiang Nie.)
\IEEEcompsocthanksitem Yangyang Guo and Liqiang Nie are with
Shandong University, China. E-mail: \{guoyang.eric, nieliqiang\}@gmail.com.
\IEEEcompsocthanksitem Zhiyong Cheng is with Shandong Artificial Intelligence Institute, Qilu University of Technology (Shandong Academy of Sciences), China.
E-mail: jason.zy.cheng@gmail.com.
\IEEEcompsocthanksitem Qi Tian is with Huawei Noah's Ark Laboratory, Huawei, China. Email: tian.qi1@huawei.com.
\IEEEcompsocthanksitem Min Zhang is with Harbin Institute of Technology, Shenzhen, China. Email: zhangminmt@hotmail.com.

}}

%



\maketitle

\begin{abstract}
Recent studies have pointed out that many well-developed Visual Question Answering (VQA) models are heavily affected by the language prior problem. It refers to making predictions based on the co-occurrence pattern between textual questions and answers instead of reasoning upon visual contents. To tackle this problem, most existing methods focus on strengthening the visual feature learning capability to reduce this text shortcut influence on model decisions. However, few efforts have been devoted to analyzing its inherent cause and providing an explicit interpretation. It thus lacks a good guidance for the research community to move forward in a purposeful way, resulting in model construction perplexity towards overcoming this non-trivial problem. In this paper, we propose to interpret the language prior problem in VQA from a class-imbalance view. Concretely, we design a novel interpretation scheme whereby the loss of mis-predicted frequent and sparse answers from the same question type is distinctly exhibited during the late training phase. It explicitly reveals why the VQA model tends to produce a frequent yet obviously wrong answer, to a given question whose right answer is sparse in the training set. Based upon this observation, we further propose a novel loss re-scaling approach to assign different weights to each answer according to the training data statistics for estimating the final loss. We apply our approach into six strong baselines and the experimental results on two VQA-CP benchmark datasets evidently demonstrate its effectiveness. In addition, we also justify the validity of the class imbalance interpretation scheme on other computer vision tasks, such as face recognition and image classification.
\end{abstract}


\section{Introduction}\label{introduction}
\IEEEPARstart{V}{i}sion and language are two ubiquitous elements in human cognition. One key artificial intelligence effort in bridging these two is to answer natural language questions about a visual scene, the \emph{de facto} Visual Question Answering (VQA) task. VQA has gained great progress over the past few years owing to the considerable development of computer vision and natural language processing. Conventional methods generally cast it as a \textbf{classification problem}, where the image and question are respectively handled via Convolutional Neural Networks (CNNs) and Recurrent Neural Networks (RNNs), and the fused multi-modal features are classified into candidate answers~\cite{vqa1, updown}.

\begin{figure}
  \centering
  \includegraphics[width=0.9\linewidth]{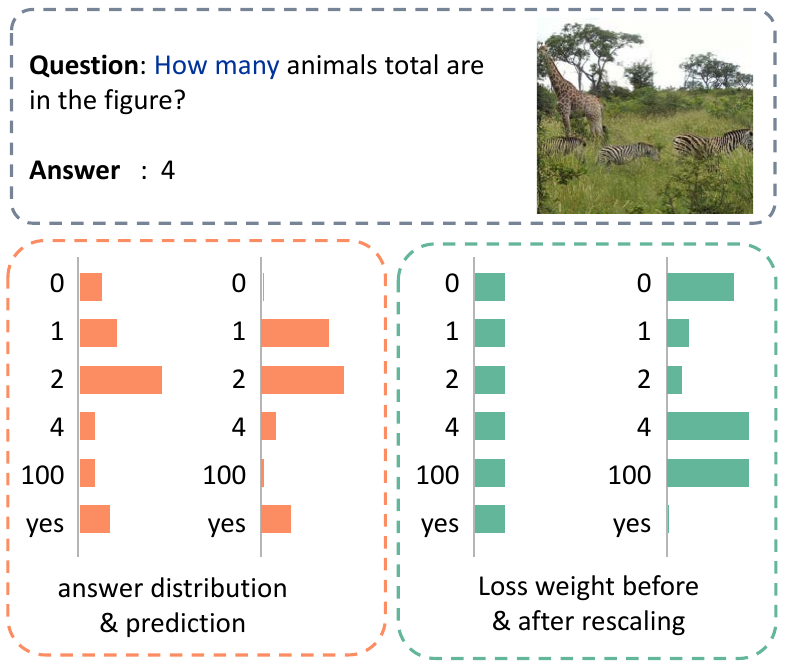}
  \caption{Illustration of why the language prior problem arises and our solution. On the left orange bars, 1) the answer distribution in the training set for the current question type \emph{how many} is biased, which leads the answer prediction to drift towards frequent answers including an out-of-type one - \emph{yes}. On the right green bars, 2) the original loss weight is evenly distributed over each candidate answer, while our solution is to re-scale it for the model decision balancing.}\label{fig:example}
\end{figure}

Though a lot of VQA approaches have achieved great success over various benchmarks~\cite{vqa1, vqa2}, one notorious issue is that they are often hindered by the widely investigated language prior problem. That is, these models answer questions without comprehensively exploiting the visual contents. In fact, existing VQA methods are often brittle and easy to be fooled by some statistical shortcuts owing to the linguistically superficial correlation between questions and answers. For example, the \emph{how many}\footnote{We refer \emph{question type} to the first few words in the given question.} questions are frequently be answered with answer \emph{2} in the training set~\cite{vqa1}. This simple correlation misleads models to overwhelmingly reply to \emph{how many} questions with \emph{2} yet not truly reasoning the given image (see the training answer distribution and prediction in Figure~\ref{fig:example} for an example). As a result, it is problematic to assert that the previous benchmark progress is driven by the faithful visual understanding rather than merely learning the dataset bias. To tackle this, a diagnostic benchmark, i.e., VQA-CP (VQA under Changing Priors)~\cite{vqacp} has recently been constructed. It re-splits VQA v1~\cite{vqa1} and VQA v2~\cite{vqa2} datasets in a way such that the distribution of answers per question type is significantly distinct between training and testing sets. As can be expected, the performance of many prior state-of-the-art VQA models~\cite{hierarchical, san, updown} largely degrades on this dataset.

It is worth mentioning that great success has been achieved by prior efforts pertaining to overcoming the language prior problem in VQA. These methods can be roughly grouped into two categories: \textit{single-branch} and \textit{two-branch}. Specifically, approaches in the first category directly strengthen the visual feature learning for weakening the textual shortcut influence on model decision~\cite{hint, supervise, grounding, critical}. For instance, HINT~\cite{hint} and SCR~\cite{critical} leverage additional visual annotations to align the image region-level importance\footnote{The image region importance is expressed via the model gradient magnitude of the ground-truth answers, i.e., Grad-CAM~\cite{grad-cam}. with human attention map}. Though yielding competitive improvement over their rivals, nonetheless, collecting such human annotations is expensive and burdensome, limiting their practicality in adaptation across datasets. By contrast, the two-branch methods deliver dominated performance due to their explicit counterwork over the language priors~\cite{adversarial-nips, css, rubi, lmh}. A typical design is to append an additional question-only training branch to the orthodox question-image one, which intentionally captures the language priors and is further suppressed by the latter branch. The computational cost from this category of methods is thereby increased since another new question-only branch is introduced to the backbone model.

In general, the pipeline of the existing studies can be summarized into two consecutive steps: 1) describing the background and consequence of the language prior problem in VQA, and 2) developing effective methods to mitigate it and demonstrating the superiority over their contenders. However, little attention has been paid to interpreting why it drives models to blindly answer questions even though the corresponding image shows very intuitive cues for the right answer. Without an interpretation, the SOTA results on the VQA-CP benchmark mostly manifest to be skeptical. In other words, it is dubious that the performance improvement is from the alleviation of the language priors, or the over-fitting on the new curated datasets. Moreover, another imperative issue of the existing approaches is that they often involve many parameters to the backbone model, resulting in heavier computations. Taking the more advanced two-branch approaches for example, the model at least increases the parameters for the answer prediction of the question-only branch (e.g., proportional to the product of the candidate answer size and the number of question features), which is burdensome and demands further optimization.

To deal with the above two issues, in this paper, we propose to bridge the connection between the language prior in VQA and the class-imbalance problem, and make contributions from the following two aspects. Firstly, to analyze the inherent reasons that cause the language prior problem in VQA models, we offer to revisit the problem from a class-imbalance view and contribute a novel viewpoint to interpret it. Recall that each candidate answer is taken as a single class and the answer distribution under each question type is highly biased. This observation is consistent with that in the class-imbalance classification scenario when restricting the same question type for each instance. In the light of this, we argue that the language prior problem is essentially aroused by the imbalanced answer class distribution per question type. Accordingly, the VQA models are severely affected by the direct correlation between the question type and its frequent answers, leaving the visual features overlooked for answer prediction. To validate this point, we make two assumptions: 1) the loss values are diverse from the trained models pertaining to answers with different numbers in the training set; and 2) the gradient norm of the parameters from the question encoding layer is much larger than that of the classification layer. The former assumption interprets why the model cannot fit the tail classes (sparse answers per question type) even with more training iterations. And the latter one points out that the parameters from the question encoding layer should be the blame for the language prior learning. This seamlessly matches the application of previous two-branch approaches~\cite{adversarial-nips, rubi} that overcomes this problem through tuning the question encoding parameters, and supports  these methods. After validating both assumptions, we then propose a loss re-scaling approach, which borrows the idea from class-imbalance solutions. The key of our approach is to assign a distinct weight to each answer for estimating the instance loss based on its associated question type. Specifically, smaller weights are assigned to frequent answers while larger weights are attached to sparse ones. Notably, the proposed method is model-agnostic, which means it can be plugged into most existing methods to overcome the language prior problem. Besides, the loss re-scaling introduces \textbf{no} training parameters, i.e., \textbf{zero} incremental inference time to the backbone model. Compared to previous methods, the increased computational cost is thus negligible.

To testify the effectiveness of the proposed loss re-scaling approach, we conduct extensive experiments with six backbone methods on the VQA-CP v2 and VQA-CP v1 datasets. The experimental results demonstrate the enhanced performance when applying our method to the backbone models under various settings. Moreover, we also evaluate the viability of the class-imbalance interpretation scheme on other Computer Vision (CV) tasks which are also hindered by the class-imbalance problem.

In summary, the contributions of this paper are three-fold:
\begin{itemize}
  \item We revisit the language prior problem in VQA from a class-imbalance view. The proposed interpretation scheme is practical and can be extended to other relevant CV fields. To the best of our knowledge, this is the first work to provide interpretations for the cause of the language prior problem in VQA.
  \item Based on this novel viewpoint, we further design a simple yet effective loss re-scaling method to alleviate the language prior effect in VQA.
  \item Extensive experiments are performed on several benchmark datasets and promising results from the proposed method can be observed. In addition, the code has been released to facilitate other researchers\footnote{\url{https://github.com/guoyang9/class-imbalance-VQA}.}.
\end{itemize}

\section{Related Work}\label{related_work}
In this section, we discuss three directions of related literature which are highly relevant to this work: Visual Question Answering, Language Prior Problem in VQA and Data Imbalance Issue.
\subsection{Visual Question Answering}
Visual Question Answering constitutes a typical vision-and-language task\cite{tip-vqa1, tip-vqa2, tip-caption, hengtao1, hengtao2}, involving the inputs of an image and a natural language question. Almost all the existing VQA models follow such a training paradigm: the image and the question are encoded with CNNs and RNNs, respectively, followed by a multi-modal fusion module~\cite{mm} and a classifier to predict the right answer(s). Based on this paradigm, current approaches can be roughly classified into four categories~\cite{survey}: \emph{Joint Embedding}, \emph{Attention Mechanism-based}, \emph{Compositional Models}, and \emph{Knowledge-enhanced}. The \emph{Joint Embedding} methods~\cite{vqa1, ask} present to encode the image and question into a common latent space, in which the visual reasoning is performed.  Later on, the \emph{attention mechanism} is introduced to focus on salient image regions~\cite{san, strong} or with question words~\cite{hierarchical, mfb}. The modular structure of questions is also exploited for more explicit reasoning of \emph{Compositional VQA models}~\cite{nmn}. Lastly, it is natural that some questions cannot be directly answered from the image information merely, requiring accessing external knowledge for accurate prediction. \emph{Knowledge Enhanced} approaches~\cite{askme, fvqa} thus integrate  knowledge bases such as DBpedia~\cite{dbpedia} or ConceptNet~\cite{conceptnet} into VQA.

Beyond the simple reasoning on images, researchers have also explored other auxiliary information and built more challenging benchmarks. For instance, Visual Commonsense Reasoning (VCR) requires a VQA model to provide an accurate answer as well a rationale (expressed via text) explaining why the answer is chosen. TextVQA asks models to read and reason the scene text in images. And video question answering~\cite{videoqa1, videoqa2, videoqa3} extends the question answering procedure from images to videos. 

\subsection{Language Prior Problem in VQA}
A large body of studies pointed out that many VQA models are greatly affected by the detrimental language prior problem~\cite{vqacp, adversarial-nips, lpscore, questiontype, unshuffle}. To address it, research have devoted efforts into the following two sides.

\textbf{From the data side.} It is extremely challenging to circumvent the dataset bias when crowd-sourcing with human annotators. In VQA v1~\cite{vqa1}, Antol et al. collected the first large-scale VQA dataset, which manifests a lot of `visual priming biases'. For example, when people ask the question `Is there a clock tower in the picture?', there actually contain clock towers in the given image. This poses negative effects on the VQA model learning since a balanced dataset should at least involves images without clock towers for this kind of questions. To amend this problem, ~\cite{yinyang, vqa2} later presented the VQA v2 dataset which introduces complementary samples in such a way that each question is associated with a pair of images yet different answers. However, ~\cite{vqacp, lpscore} indicate that this dataset is still limited by the bias problem, which can potentially drive VQA models to learn some language priors. In view of this, VQA-CP~\cite{vqacp} was created by re-splitting datasets following the rule that the answer distribution of training and testing sets is distinct with respect to question types. The performance of many VQA models drops dramatically on the VQA-CP datasets. Different from~\cite{vqacp} re-organizing the existing biased datasets, Johnson et al.~\cite{clevr} curated a new diagnostic 3D shape dataset, to control the question-conditional bias via rejection sampling within families of related questions. The answer distribution is therefore balanced.

\textbf{From the model side.} Studies in this side make efforts to directly designing mechanisms for reducing the language prior effect, which can be summarized into two groups: single-branch and two-branch models. Specifically, single-branch methods are proposed to directly enhance the visual feature learning in VQA. For example, HINT~\cite{hint} and SCR~\cite{critical} align the image region-level importance with the additional human attention maps. VGQE~\cite{vgqe} considers the visual and textual modalities equally when encoding the question, where the question features include not only the linguistic information from the question but also the visual information from the image. Zhu et al.~\cite{supervise} proposed a self-supervised framework to generate balanced question-image pairs. Differently, two-branch methods intend to explicitly counter the language prior learned by the model. A common processing is to introduce a question-only branch for deliberately capturing the language priors, followed by the question-image branch training to restrain it. For instance, Q-Adv~\cite{adversarial-nips} trains the above two models in an adversarial way, which minimizes the loss of the question-image model while maximizing the question-only one. More recent fusion-based methods~\cite{lmh, rubi, css} employ the late fusion strategy to combine the two answer predictions and guide the model to focus more on the answers which cannot be correctly addressed by the question-only branch.

Though the language prior problem in VQA has been extensively studied in recent years, nevertheless, few work contributes to analyzing its inherent cause, i.e., what arouses this problem is largely untapped in literature. Hence, in this work, we tentatively proposed to interpret the problem from a class-imbalance view, and developed a simple yet effective approach to tackle it.

\subsection{Data Imbalance Issue} \label{rw:data}
The data imbalance issue has been long recognized as a troublesome problem in a variety of research fields. Recent efforts mainly pursue solutions along three directions~\cite{decoupling}: 1) \emph{Data re-weighting} methods attach different weights to classes for estimating the final loss. A common solution is to assign large weights for tail classes (i.e., less frequent ones) and small weights for head classes (i.e., more frequent ones). For example, Focal loss~\cite{focal} is proposed to automatically put more focus on hard and mis-classified samples. 2) \emph{Data re-sampling} approaches target at re-sampling the highly biased data to attain a more balanced  distribution. In particular, over-sampling methods~\cite{oversample} simply repeat the minority classes several times to alleviate the training bias, while under-sampling ones~\cite{undersample} randomly choose a subset of the majority classes. Nevertheless, the two sampling methods all have some drawbacks: over-sampling is easy to over-fit and under-sampling undermines the model generalization capability.  And 3) \emph{Transfer learning}. Different from the above two lines of efforts, methods along this line aim to transfer the knowledge learned from abundant head classes to the tail ones~\cite{tip-transfer}. The knowledge can be intra-class variance~\cite{transfer1} or deep semantic features~\cite{transfer2}.

\section{Revisiting the Language Prior Problem}\label{preliminary}
\begin{table}
  \centering
  \caption{Toy example of why the language prior problem occurs under the question type \emph{how many}.}\label{tab:definition}
  \begin{tabular}{|c|c|c|c|c|}
    \hline
    \thead{ground \\ truth} & \thead{proportion in \\ training set} & prediction    & \thead{loss \\ value} & \thead{endowed \\ name}   \\
    \hline
    2                       & 80\%                                  & 4             & large                     & hard mistake              \\
    \hline
    4                       & 4\%                                   & 2             & small                     & easy mistake              \\
    \hline
  \end{tabular}
\end{table}
The language prior problem in VQA has been extensively studied in recent years, however, few work tries to interpret it from the angle of model learning, which makes the advancement on benchmarks unreliable to some extent. In light of this, we intend to fill this gap by revisiting it from a class-imbalance view. The reason is that current VQA models often treat VQA as a classification problem and the answer distribution per question type is highly imbalanced. In the following, we start with two assumptions and then empirically justify them. We further extend this interpretation scheme to two other CV tasks.

\subsection{New Viewpoint}
We firstly present a novel viewpoint for the class-imbalance issue according to the loss computation.

\begin{figure}
  \centering
  \subfloat[question type - \emph{how many}]{
  \centering
  \includegraphics[width=0.48\linewidth]{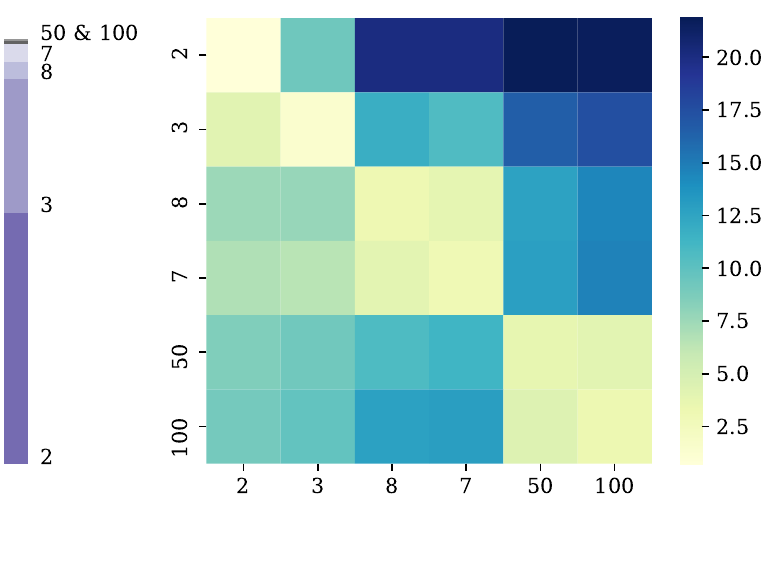}\label{fig:conf-vqa-1}
  }
  \subfloat[question type - \emph{what color}]{
  \centering
  \includegraphics[width=0.48\linewidth]{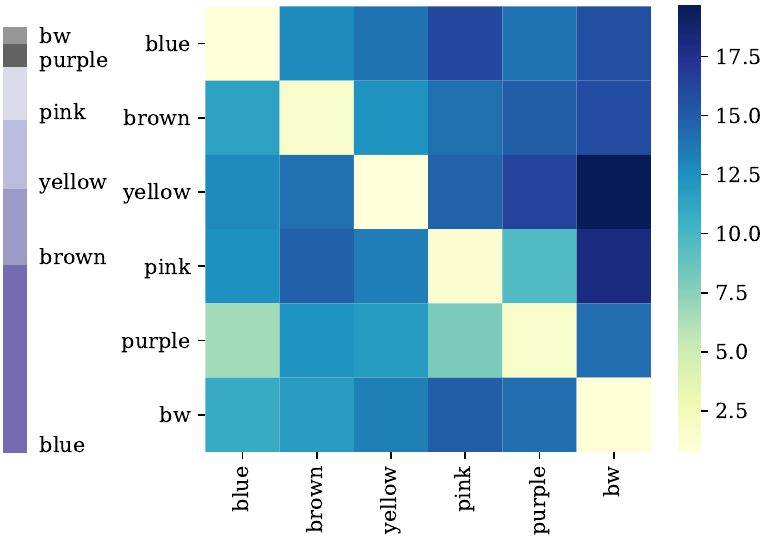}\label{fig:conf-vqa-2}
  }
  \caption{Answer distribution and loss values from two biased question types \emph{how many} on the left and \emph{what color} on the right. The bar figure corresponds to the proportion of each answer. Regarding the confusion matrix, the x-axis denotes the predicted answers from frequent to sparse, y-axis represents the ground-truth answers with the same order, and value signifies the loss. Note that the numbers of each case are kept identical for fair analysis.}\label{fig:conf_vqa}
\end{figure}

\begin{assumption}
The loss value of hard mistake samples is larger than that of easy mistake ones during the late training phase (when the learning converges).
\end{assumption}

We use a toy example in Table~\ref{tab:definition} to illustrate this assumption\footnote{Note that we only consider those instances which are mis-predicted, since the loss from the correctly predicted ones would be zero and back-propagate no gradients.}. As shown in Table~\ref{tab:definition}, \emph{2} takes 80\% of the answers to questions under the question type \emph{how many}, while \emph{4} accounts for only 4\% in the training set. After the model is trained and the learning goes convergence, we assume that the loss value of these instances which are incorrectly predicted to \emph{4} from \emph{2} (hard mistake) is much larger than that of the ones predicted to \emph{2} from \emph{4} (easy mistake) - the name is given according to the relative number of true and predicted answers. Specifically, during the early training phase, the loss of both easy and hard mistakes is numerically similar. However, when it comes to the late training phase or the learning converges, the study on how the class imbalance issue affects the model prediction is particularly sparse in literature to our knowledge. In fact, there demands a deeper analysis on whether the loss computation is consistent between the early and late training phases. It is because that if the model learning is special during late training, it can provide us with insights on how the class-imbalance problem affects the model prediction when testing, and enlighten us to design directed methods.

Towards this end, we would like to empirically study it in this work. Particularly, we leverage a trained model, e.g., UpDn~\cite{updown}, to observe the loss value with respect to different mistakes. To begin with, we fix the model parameters when the learning goes convergence. Thereafter, we reuse the instances in the training set and save the resultant prediction. The statistics and results are demonstrated in Figure~\ref{fig:conf_vqa}. It can be observed that the loss values from the confusion matrix in the upper triangle are much larger than that of the lower one. Especially for \emph{how many} questions, the values predicted from sparse answers (i.e., \emph{50}, \emph{100}) to the most frequent one \emph{2} are the highest among others, denoting that the model penalizes these hard mistakes to a great extent. In contrast, the loss values of easy mistakes (corresponding to the lower triangle) tend to be much smaller, signifying that the model seems to overlook them during the late training. This finding is quite beneficial to understand the class-imbalance problem and we will show its validity in other domains in Section~\ref{cv}.

Based on the first assumption, we further analyze the overall loss values and gradient norms with respect to training iterations. The second assumption is given as:
\begin{assumption}
The gradient norm from the parameters of the question encoding layer is larger than that of the prediction layer.
\end{assumption}

\begin{figure}
  \centering
  \includegraphics[width=1.0\linewidth]{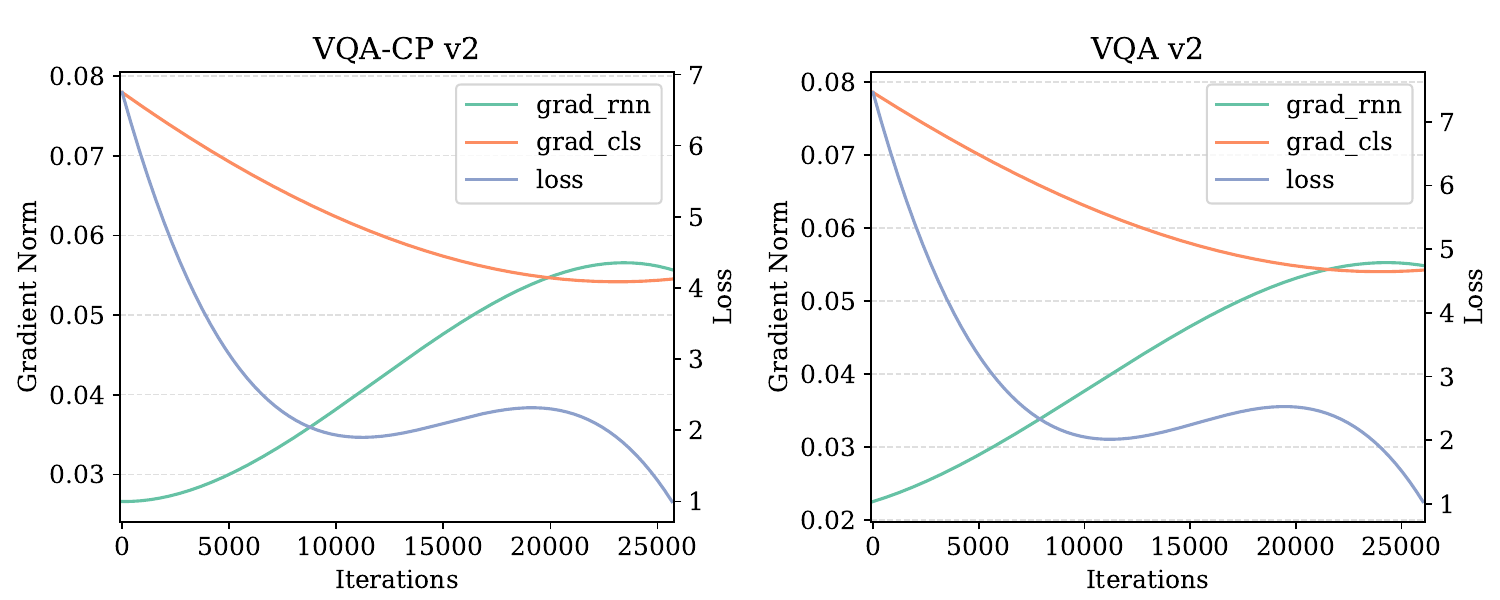}
  \caption{Gradient norm of two groups of parameters with respect to training iterations. \emph{grad\_rnn} denotes the gradient norm from the question encoding layer parameters, while \emph{grad\_cls} corresponds to the prediction layer.}\label{fig:norm_step}
\end{figure}

It is widely accepted that parameters with larger gradient norm often go with more updating when performing back-propagation. This assumption indicates that the question encoding layer affects more to the language prior learning than the prediction layer. To validate this, we plot the gradient norm from two groups of parameters with respect to the training iterations in Figure~\ref{fig:norm_step}. It can be seen that the gradient norm of the parameters from the question encoding layer continues to increase, while the parameters from the prediction layer decreases a lot with more iterations. Note that the VQA models are affected more by the language prior problem in the late training phase. As a result, the parameters fluctuate 
sharper should contribute more to the learning of language priors. Namely, the question encoding layer should be the blame for the language prior problem, rather than the prediction one. This observation is consistent with most two-branch studies~\cite{adversarial-nips, rubi}, which aim to tune the question encoder for reducing the language priors. And this is the first time to evidently interpret why we should follow such a paradigm to alleviate this problem. 

\subsection{Extension to other CV Tasks}\label{cv}
\begin{figure}
    \centering
    \subfloat[Face Recognition]{
    \centering
    \includegraphics[width=0.48\linewidth]{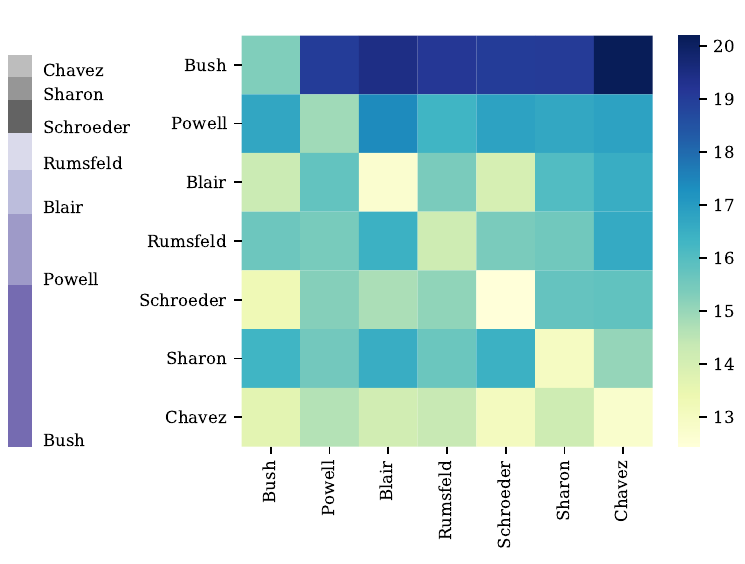}\label{fig:conf-image-1}
    }
    \subfloat[Image Classification]{
    \centering
    \includegraphics[width=0.48\linewidth]{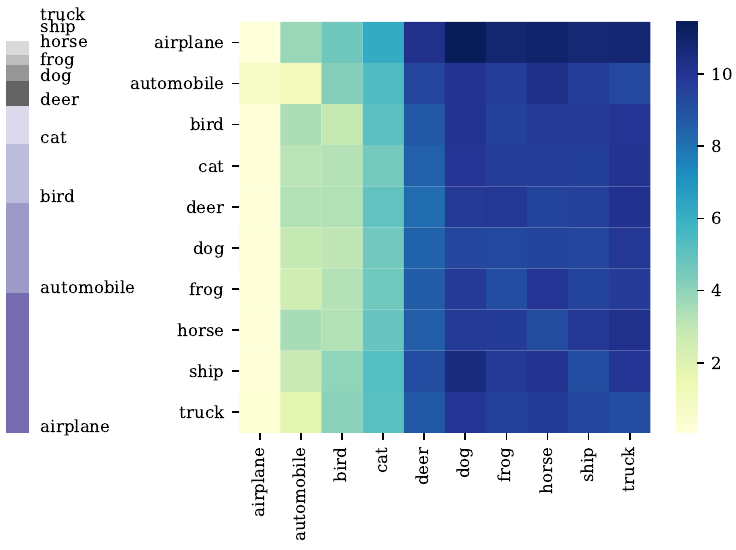}\label{fig:conf-image-2}
    }
    \caption{Class distribution and loss values from two CV tasks. The left one corresponds to seven persons from the LFW face recognition dataset. And the right one denotes the ten classes in the imbalanced CIFAR-10 dataset.}
    \label{fig:conf_image}
\end{figure}

To explore whether the first class-imbalance interpretation scheme works in other highly imbalanced CV scenarios, we additionally tested the face recognition and image classification tasks and show the results in Figure~\ref{fig:conf_image}. The method and dataset employed for face recognition are respectively the Center Loss~\cite{face} the LFW dataset (Labeled Face in the Wild)~\cite{lfw}, where we sampled seven persons with distinctive numbers in the training set. While for the image classification, we leveraged the imbalanced CIFAR-10 dataset~\cite{cifar} and the Resnet~\cite{resnet} Network for evaluation. From Figure~\ref{fig:conf_image}, we can observe that for both tasks, the results show a similar manifestation with the ones in Figure~\ref{fig:conf_vqa}. For example, when incorrectly predicting from the sparse \emph{Chavez} class to other more frequent ones in face recognition, i.e., hard mistakes, the loss values become much larger than others. This can draw the parameters to avoid these mistakes and predict to other classes albeit yielding incorrect predictions. Similar observations can also be found on the image classification task. Another interesting point from the imbalanced image classification one is that the model often ignores the mistakes from more frequent classes while penalizes the ones from sparser ones.

\section{Proposed Method}\label{model}
In this section, we present our method to overcome the language prior problem in VQA. Our method borrows the idea from the approaches attacking the class-imbalance issue in other domains. Specifically, in this work, we adopt the loss re-weighting approach to deal with this problem. In particular, we design a novel mechanism to assign different weights over different answers (classes) under each question type to estimate the final loss. Before delving into the details of the loss re-scaling method, we firstly recap the basic learning functions in VQA.

\subsection{Preliminary}\label{model:solution}
The goal of VQA is to provide an accurate answer $\hat{a}$ to a given question $Q$ upon an image $I$, which can be achieved by,
\begin{equation}\label{equ:goal}
  \hat{a} = \arg \max_{a \in A} p(a | Q, I; \Theta),
\end{equation}
where $\Theta$ denotes the model parameters and $A$ represents all the candidate answers. Existing methods mostly tackle it as a classification task. There are actually two available solutions to predict the correct answer(s): Sigmoid activation function followed by Binary Cross Entropy loss and Softmax activation function followed by the Cross Entropy loss. We use Sigm-BCE and Soft-CE to denote the two solutions, respectively. It is worth noting the ground-truth for each image-question pair is not necessarily unique, that is, one question is answered by ten annotators and then the label is softened by ten. In this way, for Sigm-CE, taking each candidate answer as a single label for the binary classification problem\footnote{Notice that we omit the mini-batch summation in all loss functions for simplicity.}, is given as,
\begin{equation}\label{equ:bce}
  \begin{aligned}
    \bm{p} &= \bm{W}_p \bm{h}_f + \bm{b}_p, \\
    \hat{\bm{p}} &= \sigma (\bm{p}), \\
  \mathcal{L}_{bce} &= -  \sum_{i = 1}^{|A|} a_i \log \hat{p}_i + (1 - a_i) \log (1 - \hat{p}_i),
  \end{aligned}
\end{equation}
where $\bm{h}_f$ is the fused multi-modal feature, $\sigma (\cdot)$ represents the Sigmoid activation function, $a_i$ denotes the ground-truth answer label, $\bm{W}_p$ and $\bm{b}_p$ are the learnable matrix and bias, respectively. 

In contrast, Soft-CE tackles the multi-label classification in VQA via,
\begin{equation}\label{equ:ce}
  \begin{aligned}
    \bm{p} &= \bm{W}_p \bm{h}_f + \bm{b}_p, \\
    \hat{\bm{p}} &= \text{Softmax} (\bm{p}), \\
  \mathcal{L}_{ce} &= - \sum_{i = 1}^{|A|} a_i \log \hat{p}_i,
  \end{aligned}
\end{equation}
where $\text{Softmax} (\cdot)$ implies the Softmax activation function.

Based on the aforementioned two solutions, we propose to assign distinctive weights for each answer when computing the final loss. To this end, in the following, we firstly design an \emph{Answer Mask} module to constrain the predicted answers fall into the correct set pertaining to each question type. We then employ a novel strategy to compute the answer weights for the loss estimation.


\begin{figure}
    \centering
    \subfloat[Distribution from VQA-CP v2]{
    \centering
    \includegraphics[width=0.48\linewidth]{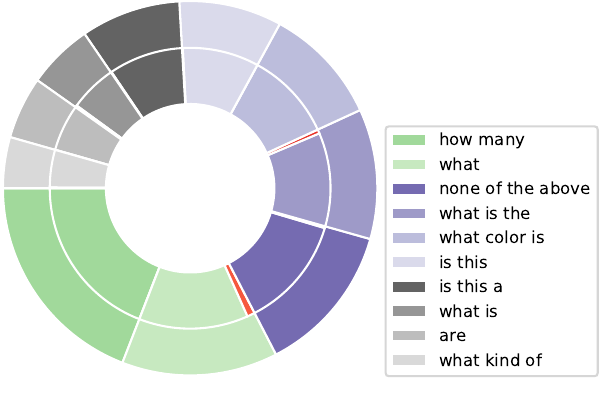}\label{fig:outlier-1}
    }
    \subfloat[Distribution from VQA v2]{
    \centering
    \includegraphics[width=0.48\linewidth]{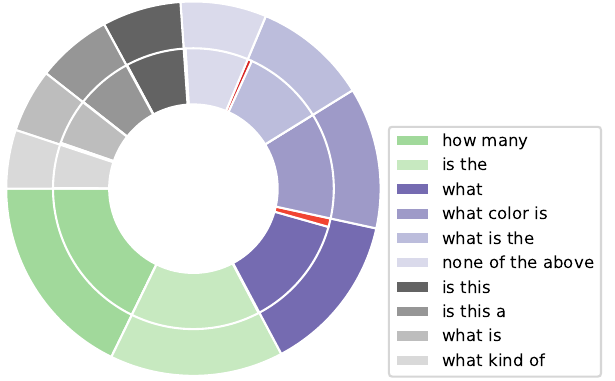}\label{fig:outlier-2}
    }
    \caption{Answer distribution in the training set (outer circle) and the validation set (inner circle) of the most frequent question types. The red samples represent the answers which are out of the corresponding answer set in the training set. }
    \label{fig:outlier}
\end{figure}

\begin{figure*}
  \centering
  \subfloat[Backbone learning pipeline]{
  \includegraphics[width=0.36\linewidth]{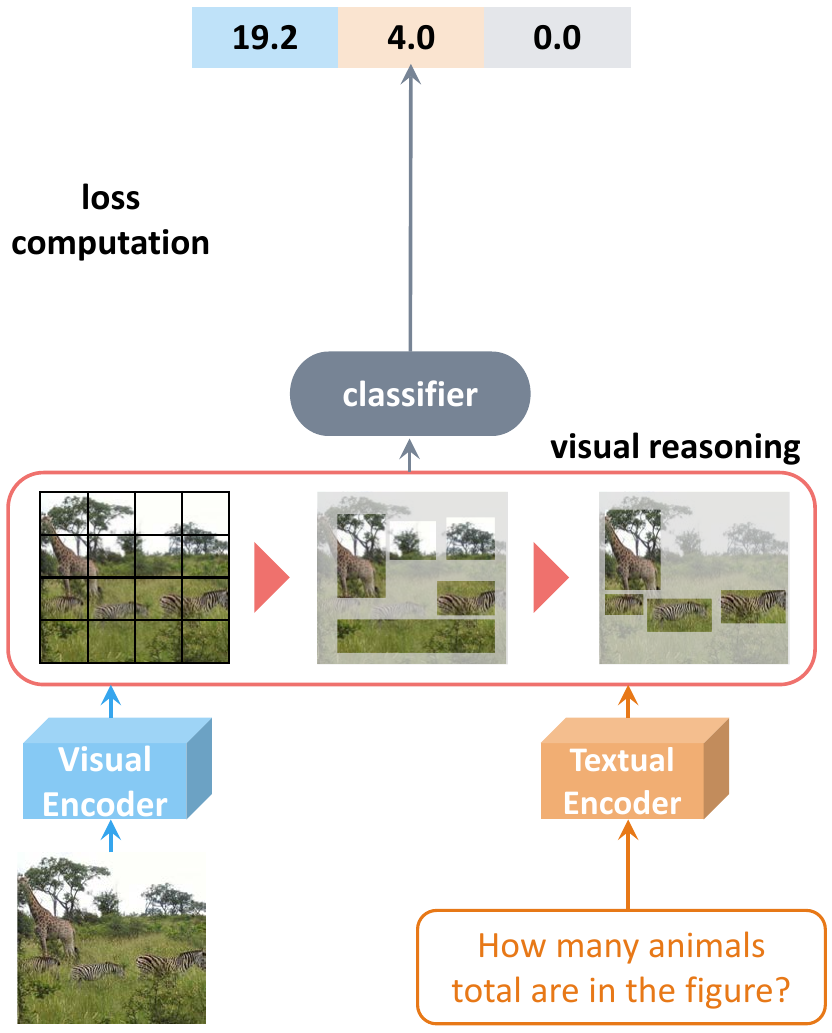}\label{fig:baseline}
  }
  \hfill
  \subfloat[Our proposed learning pipeline]{
  \includegraphics[width=0.50\linewidth]{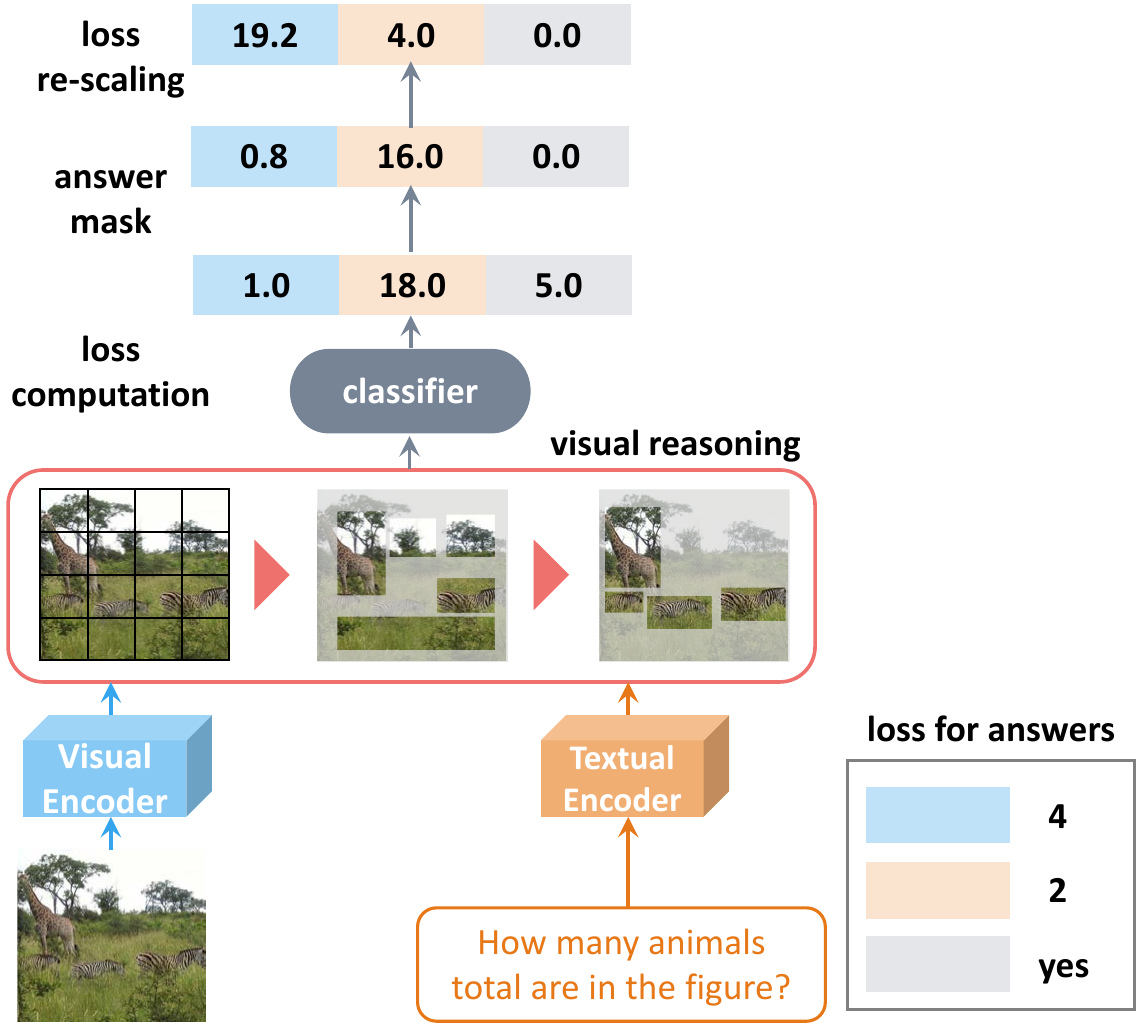}\label{fig:model}
  }
  \caption{Visual comparison of loss estimation between the backbone learning pipeline and ours. We use three answers, i.e., \emph{4}, \emph{2} and \emph{yes}, to intuitively illustrate how the proposed two modules work.}\label{fig:pipeline}
\end{figure*}

\subsection{Answer Mask}
Let's revisit the example in Figure~\ref{fig:example}, for the question type \emph{how many}, there even exists an out-of-type answer \emph{yes}. This perplexes a lot as very rare \emph{how many} questions should be answered with binary answers \emph{yes} or \emph{no}. One possible reason is that the \emph{yes} answer takes large parts of the answers, which makes the model unexpectedly drift to this frequent label. To overcome this problem and ensure the predicted answers lie into the specific answer set of the given question type, we introduce an answer mask in this section. The motivation is supported by the findings in Figure~\ref{fig:outlier}, where only a very small number of instances in the validation set are distributed to the out-of-type answers from the corresponding answer set. Concretely, the answer mask module takes the question feature as inputs, and then employs a fully-connected layer equipped with a non-linear activation function,
\begin{equation}\label{equ:mask}
  \bm{m}_p = f(\bm{W}_q \bm{q}),
\end{equation}
where the size of $\bm{m}_p \in \mathbb{R}^{|A|}$ is the same with the candidate answer size, $f(x)$ denotes the activation function, and $\bm{W}_q$ are the learnable weights. To enable the answer mask modeling, we collect the true mask label $\bm{m}_a \in \mathbb{R}^{|A|}$ where the labels belonging to the given question type are set to 1 and others are 0. In this way, the BCE loss is used to train this module,
\begin{equation}\label{equ:loss_mask}
  \mathcal{L}_{mask} = \sum_{i = 1}^{|A|} m_a^i \log m_p^i + (1 - m_a^i) \log (1 - m_p^i),
\end{equation}
where $m_a^i$ and $m_p^i$ represent the $i$-th element in $\bm{m}_a$ and $\bm{m}_p$, respectively.

In fact, a straightforward design of $f(x)$ is to adopt the sigmoid activation function, which can map the features to a mask value within $(0, 1)$. However, we favor the output to be larger than 0.5 in most times so that the final predicted answer would be grounded more by this mask. Note that deploying all-ones mask will lead to over-fitting and is not flexible. We therefore devise a generalized softplus function in Equation~\ref{equ:softplus},
\begin{equation}\label{equ:softplus}
  f(x) = \max (1, \frac{1}{\alpha} \log (1 + \exp \alpha x)),
\end{equation}
where $\alpha$ is a hyper-parameter which is fixed to 1 in our experiment.

\subsection{Loss Re-scaling}
The key to our loss re-scaling approach is to attach distinctive weight to each answer. To achieve this, we intuitively introduce a fixed weight, where it can prevent the hard mistakes to update the parameters harshly while guide the easy mistakes for more influential model learning. In particular, the weight is obtained via,
\begin{equation}\label{equ:miu}
  \mu_i = \frac{\sum_{k}^{|A|}n_k^j - n_i^j}{n_i^j},
\end{equation}
where $n_i^j$ represents the number of answer $a_i$ under the question type $qt_j$ (for other answers not in the answer set of the current question type, $\mu_i$ is equal to 1). Under this situation, recollect the previous example in Table~\ref{tab:definition}, the loss weights for answer `2' and `4' are $(1-0.8) / 0.8 = 0.25$, and $(1- 0.04) / 0.04 = 24$, respectively. In this way, the easy mistakes become relatively hard to propagate information, while hard mistakes gain more chance to update parameters, especially for earlier layers. In implementation, we found that large weights made the training more unstable, we therefore employ a smoothing function and truncate too large values to be 100,
\begin{equation}\label{equ:miu_soft}
  \mu_i = \max (100, \log (1 + \exp \mu_i)).
\end{equation}

This approach is specially designed for avoiding VQA models to learn language priors when training. During testing, the loss re-scaling is deactivated and only the backbone model remains. This assures that there is \textbf{no incremental computation} at all when evaluation compared with the backbone model\footnote{By zero incremental inference time, we explicitly refer to the loss re-scaling module excluding the answer mask one, since the loss re-scaling module produces a primary effect as shown in Table~\ref{tab:ablation}.}. In addition, we provide the losses and partial derivatives in Appendix~\ref{appendix}.

\subsection{Method Application}
Most mainstream VQA architectures are compatible with our method. To show how the above two modules work, we leverage a general backbone and illustrate its learning pipeline and ours in Figure~\ref{fig:pipeline}. Specifically, for each input image, it utilizes the pre-trained object detection networks (e.g., Faster R-CNN~\cite{rcnn}) to obtain the most salient objects; meanwhile, the question words are sequentially processed via a textual encoder (e.g.,  Gated Recurrent Network~\cite{gru}) to capture the semantic information. The visual reasoning is then performed, followed by the classifier for predicting  answers. As can be seen in Figure~\ref{fig:baseline}, the BCE losses for answers \emph{4}, \emph{2} and \emph{yes} are 1.0, 18.0 and 5.0, respectively. With our answer mask module in Figure~\ref{fig:model}, the \emph{yes} answer no longer takes effects since it is an out-of-type outlier for the question type \emph{how many}. Our loss re-scaling module then assigns distinct loss weights to each candidate answer according to the statistics in Table~\ref{tab:definition}. One can see that the loss for the sparse answer \emph{4} is augmented while for the frequent answer \emph{2} is diminished.

\section{Experimental Setup}\label{setup}
\begin{table*}[htbp]
  \centering
  \caption{Accuracy comparison between the proposed method and baselines over the VQA-CP v2 test and VQA v2 val sets. Regarding the method category, `Plain' represents VQA methods without specialization for overcoming the language prior problem. `Two-branch' and `Single-branch' distinguish whether there is another question-only training branch. `$-$' and `$\ddagger$' denote the number is not available and our implementation, respectively.  Symbols `$\dag$' denotes the statistical significance with two-sided t-test of $p < 0.01$, compared to the corresponding backbone. The best performance in current splits is highlighted in bold.}\label{tab:baseline1}
  \scalebox{1.0}{
  \begin{tabular}{c||l||cccc||cccc}
    \toprule
    \multirow{2}{*}{Method Category} & \multirow{2}{*}{Method}
                                                & \multicolumn{4}{c||}{VQA-CP v2 test}    & \multicolumn{4}{c}{VQA v2 val}              \\
                                                \cmidrule(lr){3-6}                      \cmidrule(lr){7-10}
                                                && Y/N   & Num.  & Other & All           & Y/N   & Num.  & Other & All                  \\
    \midrule
    \multirow{8}{*}{Plain}
    & Question-only~\cite{vqacp}                & 35.09 & 11.63 & 07.11 & 15.95         & 67.95 & 30.97 & 27.20 & 43.01                 \\
    & SAN~\cite{san}                            & 38.35 & 11.14 & 21.74 & 24.96         & 70.06 & 39.28 & 47.84 & 52.41                 \\
    & NMN~\cite{nmn}                            & 38.94 & 11.92 & 25.72 & 27.47         & 73.38 & 33.23 & 39.93 & 51.62                 \\
    & MCB~\cite{mcb}                            & 41.01 & 11.96 & 40.57 & 36.33         & 77.91 & 37.47 & 51.76 & 59.71                 \\
    & HAN~\cite{han}                            & 52.25 & 13.79 & 20.33 & 28.65         &   -   &   -   &   -   &   -                   \\
    & UpDn~\cite{updown}                        & 42.27 & 11.93 & 46.05 & 39.74         & 81.18 & 42.14 & 55.66 & 63.48                 \\
    & UpDn~\cite{updown}\dag                    & 42.31 & 11.84 & 44.35 & 38.80         & 80.59 & 40.51 & 53.69 & 62.07                 \\
    & LXMERT~\cite{lxmert}$\ddagger$            & 46.70 & 27.14 & 61.20 & 51.78 & \bf{88.34} & \bf{56.64} & \bf{65.78} & \bf{73.06}  \\
    \midrule
    \multirow{8}{*}{Two-branch}
    & AdvReg~\cite{adversarial-nips}            & 65.49 & 15.48 & 35.48 & 41.17         & 79.84 & 42.35 & 55.16 & 62.75                 \\
    & Rubi~\cite{rubi}                          & 68.65 & 20.28 & 43.18 & 47.11         &   -   &   -   &   -   & 61.16                 \\
    & LM~\cite{lmh}$\ddagger$                   & 73.65 & 41.44 & 42.82 & 51.68         & 80.37 & 40.62 & 53.53 & 61.92                 \\
    & LMH~\cite{lmh}$\ddagger$                  & 70.29 & 44.10 & 44.86 & 52.15         & 74.18 & 38.43 & 53.05 & 59.07                 \\
    & CSS~\cite{css}                            & 43.96 & 12.78 & 47.48 & 41.16         & - & - & - & -  \\
    & CSS+LMH~\cite{css}                        & \bf{84.37} & 49.42 & 48.21 & 58.95 & 73.25 & 39.77 & 55.11 & 59.91  \\
    & CSS~\cite{css}$\ddagger$                  & 40.88 & 12.32 & 45.23 & 38.95 & 80.28 & 41.90 & 53.81 & 62.20  \\
    & CSS+LMH~\cite{css}$\ddagger$              & 77.73 & 50.92 & 45.30 & 55.61 & 42.16 & 35.05 & 52.01 & 46.08  \\
    \midrule
    \multirow{4}{*}{Single-branch}
    & GVQA~\cite{vqacp}                         & 57.99 & 13.68 & 22.14 & 31.30         & 72.03 &   -   &   -   & 48.24                 \\
    & HINT~\cite{hint}                          & 67.27 & 10.61 & 45.88 & 46.73         & 81.18 & 42.99 & 55.56 & 63.38                 \\
    & SCR~\cite{critical}                       & 72.36 & 10.93 & 48.02 & 49.45         & 78.8  & 41.6  & 54.5  & 62.2                  \\
    & VGQE~\cite{vgqe}                          & 66.35 & 27.08 & 46.77 & 50.11         &   -   &   -   &   -   & 63.18                 \\
    \midrule
    \multirow{6}{*}{Ours}
    & UpDn+Ours                                 & 68.42$\dag$   & 21.71$\dag$   & 42.88$\dag$   & 47.09$\dag$   & 64.22$\dag$   & 39.61$\dag$   & 53.09$\dag$   & 55.50$\dag$ \\
    & LM+Ours                                   & 73.74$\dag$   & 45.14$\dag$   & 44.59$\dag$   & 53.17$\dag$   & 76.28$\dag$   & 36.61$\dag$   & 52.71         & 59.45$\dag$\\
    & LMH+Ours                                  & 72.82         & 48.00$\dag$   & 44.46$\dag$   & 53.26$\dag$   & 68.21$\dag$   & 36.37$\dag$   & 52.29$\dag$   & 56.81$\dag$\\
    & CSS+Ours                                  & 68.45$\dag$ & 40.79$\dag$ & 44.18 & 50.73$\dag$
                                                & 77.73$\dag$ & 41.39$\dag$ & 53.76$\dag$ & 61.14  \\
    & CSS+LMH+Ours                              & 83.95 & 47.81$\dag$ & 44.59$\dag$ & 56.55$\dag$ 
                                                & 70.52$\dag$    & 33.56$\dag$    & 50.83$\dag$    & 55.96$\dag$   \\
    & LXMERT+Ours                               & 79.77$\dag$ & \bf{59.06}$\dag$ & \bf{61.41} & \bf{66.40}$\dag$ 
                                                & 85.32$\dag$ & 52.07$\dag$ & 62.60$\dag$ & 69.76$\dag$  \\
    \bottomrule
  \end{tabular}
  }
\end{table*}

We conducted extensive experiments on three benchmark datasets to validate the effectiveness of the proposed method. In particular, the experiments are mainly performed to answer the following research questions:
\begin{itemize}
  \item \textbf{RQ1}: Can the proposed loss re-scaling method improve the state-of-the-art VQA models?
  \item \textbf{RQ2}: How do the answer mask and loss re-scaling modules perform on both solutions in Section~\ref{model:solution}?
  \item \textbf{RQ3}: Is the proposed loss re-scaling method superior than other loss re-weighting approaches?
  \item \textbf{RQ4}: Why does the proposed method outperform the backbone method?
\end{itemize}

In the following, we first provide the basic information of the evaluated benchmark datasets. We then detail the standard evaluation metric, followed by the key baselines used in this work, and we end this section with some implementation details.
\subsection{Datasets}
We evaluated our proposed method mainly on the two VQA-CP datastes: VQA-CP v2 and VQA-CP v1~\cite{vqacp}, which are well-known benchmarks for evaluating the models' capability to overcome the language prior problem. The VQA-CP v2 and VQA-CP v1 datasets consist of $\sim$122K images, $\sim$658K questions and $\sim$6.6M answers, and $\sim$122K images, $\sim$370K questions and $\sim$3.7M answers, respectively.  Moreover, the answer distribution per question type is significantly different between training and testing sets. Other than VQA-CP, we also reported results on the VQA v2, a more biased benchmark, for completeness. For all the datasets, the answers are divided into three categories: \emph{Y/N}, \emph{Num.} and \emph{Other}.
\subsection{Evaluation Metric}
We adopted the standard metric in VQA for evaluation~\cite{vqa1}. For each predicted answer $a$, the accuracy is computed as,
\begin{equation}\label{equ:metric}
  Acc = \text{min} (1, \frac{\#\text{humans that provide answer $a$}}{3}).
\end{equation}
Note that each question is answered by ten annotators, and this metric takes the human disagreement into consideration~\cite{vqa1, vqa2}.

\subsection{Tested Backbones}
We mainly tested our method over the following six backbones. Thereinto, UpDn is the most popular VQA baseline, LXMERT is a recently proposed cross-modality transformer-based framework. LM, LMH, CSS and CSS+LMH are developed to intentionally alleviate the language prior problem in VQA.

\textbf{UpDn}~\cite{updown} firstly leverages the pre-trained object detection frameworks to obtain salient object features for high-level reasoning. It then employs a simple attention network to focus on the most important objects which are highly related with the given question.

\textbf{LXMERT}~\cite{lxmert} is built upon the Transformer encoders to learn the connections between vision and language. It is pre-trained with diverse pre-training tasks on a large-scale dataset of image and sentence pairs. It is worth noting that we did not apply the answer mask module to LXMERT to avoiding undermining its pre-trained linguistic features.

\textbf{LM}~\cite{lmh} is a typical two-branch method which takes UpDn as baseline and combine the answer prediction from the two branches (i.e., question-image and question-only) in an ensemble mode.

\textbf{LMH}~\cite{lmh} extends LM by introducing another entropy loss to encourage the question-only branch prediction to be non-uniform, posing a greater impact on the ensemble.

\textbf{CSS}~\cite{css} generates counterfactual training samples by masking critical image objects or question words. These samples are then trained with negative labels.

\textbf{CSS+LMH}~\cite{css} combines CSS and LMH together to achieve better performance.

\subsection{Implementation Details}
We implemented our method based on the publicly available codes for all the backbones. For the answer mask module, we appended a dropout layer with the dropout rate tuned from 0.0 to 1.0 with a step size of 0.1. And the answer mask loss weight is fixed to 1.0. For the loss re-scaling approach, we first trained the answer mask module until converging, and then fine-tuned the model with the re-scaling weights for another ten epochs. Note that we kept all the other settings being unchanged, such as the learning rate, optimizer, mini-batch size. 
\section{Experimental Results}\label{results}
\begin{table}[htbp]
  \centering
  \caption{Accuracy comparison between the proposed method and baselines over the VQA-CP v1 dataset. `$-$' and `$\ddagger$' denote the number is not available and our implementation, respectively. Symbols `$\dag$' denotes the statistical significance with two-sided t-test of $p < 0.01$, compared to the corresponding backbone. The best performance in current splits is highlighted in bold.}\label{tab:baseline2}
  \scalebox{1.0}{
  \begin{tabular}{l|cccc}
    \toprule
    \multirow{2}{*}{Method}             & \multicolumn{4}{c}{VQA-CP v1 test}     \\
                                            \cmidrule(lr){2-5}
                                        & Y/N   & Num.  & Other & All             \\
    \midrule
    Question-only~\cite{vqacp}          & 35.72 & 11.07 & 08.34 & 20.16           \\
    SAN~\cite{san}                      & 35.34 & 11.34 & 24.70 & 26.88           \\
    NMN~\cite{nmn}                      & 38.85 & 11.23 & 27.88 & 29.64           \\
    MCB~\cite{mcb}                      & 37.96 & 11.80 & 39.90 & 34.39           \\
    UpDn~\cite{updown}$\ddagger$        & 43.76 & 12.49 & 42.57 & 38.02           \\
    \midrule
    GVQA~\cite{vqacp}                   & 64.72 & 11.87 & 24.86 & 39.23           \\
    AdvReg~\cite{adversarial-nips}      & 74.16 & 12.44 & 25.32 & 43.43           \\
    LM~\cite{lmh}$\ddagger$             & 75.01 & 28.86 & 42.22 & 53.59           \\
    LMH~\cite{lmh}$\ddagger$            & 76.61 & 29.05 & 43.38 & 54.76           \\
    CSS+LMH~\cite{css}                  & 85.60 & 40.57 & 44.62 & 60.95             \\
    CSS~\cite{css}$\ddagger$            & 43.15 & 12.79 & 43.88 & 38.36             \\
    CSS+LMH~\cite{css}$\ddagger$        & 83.55 & 29.79 & 43.59 & 57.86             \\
    LXMERT~\cite{lxmert}$\ddagger$      & 54.08 & 25.05 & 62.72 & 52.82             \\
    \midrule
    UpDn+Ours                           & 44.25$\dag$       & 18.04$\dag$       & 43.03$\dag$       & 39.34$\dag$ \\
    LM+Ours                             & 75.51$\dag$       & 26.93$\dag$       & 43.67$\dag$       & 54.07$\dag$ \\
    LMH+Ours                            & 78.26$\dag$       & 29.80$\dag$       & 42.76             & 55.32  \\
    CSS+Ours                            & 74.39$\dag$       & 37.02$\dag$       & 36.22$\dag$       & 52.20$\dag$ \\
    CSS+LMH+Ours                        & 84.56$\dag$       & 32.89$\dag$       & 43.86             & 58.91$\dag$ \\
    LXMERT+Ours                         & \bf{86.82}$\dag$  & \bf{45.03}$\dag$  & \bf{62.03}$\dag$  & \bf{69.47}$\dag$ \\
    \bottomrule
  \end{tabular}
  }
\end{table}
\subsection{Overall Performance Comparison (RQ1)}
Table~\ref{tab:baseline1} and ~\ref{tab:baseline2} present the performance results on VQA-CP v2 testing, VQA v2 validation and VQA-CP v1 testing sets. The key observations from these two tables are as follows:
\begin{itemize}
  \item Our methods achieve the best over most answer categories on the VQA-CP v2 and VQA-CP v1 datasets. Since these two datasets are deemed as an effective protocol for evaluating the language prior overcoming capability, we can conclude that the proposed method is capable of dealing with this issue and superior than the existing methods.
  \item For all the six backbones, with our answer mask and loss re-scaling modules, they can consistently obtain a significant performance improvement on the VQA-CP datasets. For example, in the VQA-CP v2 dataset, the overall accuracy of UpDn boosts from 38.80\% to 47.09\%, with a 8.29\% absolute gain.
  \item The performance gap between VQA v2 val and VQA-CP v2 test in Table~\ref{tab:baseline1} quantifies the bias learning level since the VQA v2 is a more biased dataset. It can be seen that the methods specially designed for tackling the language prior problem outperform the traditional methods, and our approaches perform even better, indicating less language priors are learned.
  \item For the comparison of two-branch and single-branch methods, we can observe that two-branch ones often outperform their contemporary rivals, such as AdvReg v.s. GVQA, Rubi v.s. HINT, and LMH even surpasses the succeeding method VGQE. One noticeable reason for this is that the question-only model in the two-branch methods explicitly captures the language priors and is then suppressed by the question-image model.
\end{itemize}

In addition, we further computed the Cohen's kappa to measure  the  agreement  between labels and  predictions in Table~\ref{tab:kappa}. From the results, it can been observed that our method outperforms the backbones in most cases. However, the kappa values mostly manifest to be small. One possible reason to this is that the number of classes is too large and all the three datasets are negatively affected by the bias distribution to some extent.

\begin{table*}[htbp]
  \centering
  \caption{Cohen's kappa comparison between backbone models and our methods on three benchmark datasets.}\label{tab:kappa}
  \scalebox{1.0}{
  \begin{tabular}{l||cccc||cccc||cccc}
    \toprule
    \multirow{2}{*}{Method}
                        & \multicolumn{4}{c||}{VQA-CP v2 test}   & \multicolumn{4}{c||}{VQA-CP v1 test}   & \multicolumn{4}{c}{VQA v2 val}   \\
                            \cmidrule(lr){2-5}                      \cmidrule(lr){6-9}                      \cmidrule(lr){10-13}
                        & Y/N   & Num.  & Other & All           & Y/N   & Num.  & Other & All           &  Y/N   & Num.  & Other & All        \\
    \midrule
    UpDn                & 47.86 & 6.81  & 0.80  & 4.96 & 43.01 & 4.40 & 0.53 & 8.57  & 49.96& 14.13 & 0.72 & 8.14 \\
    UpDn+Ours           & 55.61 & 11.89 & 0.78  & 5.66 & 63.90 & 38.68 & 0.75 & 14.24  & 51.46& 15.34 & 0.81 & 9.25 \\
    \midrule
    LM                  & 50.51 & 24.79 & 0.92  & 6.06 & 53.52 & 17.48 & 0.59 & 11.34  & 49.49& 13.28 & 0.73 & 8.19 \\
    LM+Ours             & 50.61 & 24.87 & 0.92 & 6.98 & 56.35 & 13.60 & 0.58 & 11.34  & 48.85& 10.56 & 0.68 & 5.91 \\
    \midrule
    LMH                 & 50.97 & 24.99 & 0.85  & 5.42 & 55.55 & 16.70 & 0.60 & 11.03  & 49.88& 11.59 & 0.67 & 7.56 \\
    LMH+Ours            & 53.27 & 26.03 & 0.76  & 5.74 & 54.70 & 19.10 & 0.59 & 11.21  & 49.60& 9.89 & 0.65 & 6.15 \\
    \midrule
    CSS                 & 47.67 & 7.02  & 0.81  & 4.90 & 41.66 & 4.54 & 0.57 & 8.57  & 50.19& 14.27 & 0.74 & 8.26 \\
    CSS+Ours            & 51.79 & 20.98 & 0.78  & 5.19 & 66.42 & 29.16 & 0.67 & 13.44  & 51.21& 14.37 & 0.76 & 8.97 \\
    \midrule
    CSS+LMH             & 50.58 & 28.62 & 0.81  & 5.19 & 59.89 & 16.18 & 0.61 & 11.00  & 45.91& 10.52 & 0.65 & 4.76 \\
    CSS+LMH+Ours        & 51.97 & 29.21 & 0.82 & 6.23 & 61.33 & 19.60 & 0.59 & 11.93  & 49.75& 10.03 & 0.64 & 7.64 \\
    \midrule
    LXMERT              & 46.32 & 8.83 & 0.83 & 4.78 & 43.22 & 6.28 & 0.59 & 8.51  & 50.03& 13.17 & 0.69 & 7.99 \\
    LXMERT+Ours         & 49.95 & 22.51 & 0.85 & 5.43 & 56.14 & 17.94 & 0.63 & 11.25  & 50.02& 11.47 & 0.62 & 7.88 \\
    \bottomrule
  \end{tabular}
  }
\end{table*}

\subsection{Ablation Study (RQ2)}
\begin{table*}
  \centering
  \caption{Effectiveness validation of the answer mask and loss re-scaling modules. `$-$' represents the number is not available, since LM and LMH directly performs on the BCE loss. And softplus-G implies the activation function in Equation~\ref{equ:softplus}.}\label{tab:ablation}
  \scalebox{1.0}{
  \begin{tabular}{l||l||cccccc}
    \toprule
    Solution                & Module                    & UpDn  & LM    & LMH   & CSS    & CSS+LMH    & LXMERT   \\
    \midrule
    \multirow{5}{*}{Sigm-BCE}
                            & baseline                  & 38.80 & 51.68 & 52.15 & 38.95  & 55.61      & 51.78  \\
                            & +answer mask (sigmoid)    & 40.94 & 51.71 & 51.56 & 38.76  & 56.12      & -  \\
                            & +answer mask (softplus-G) & 41.69 & 52.76 & 52.70 & 38.72  & 56.55      & -  \\
                            & +loss re-scaling          & 40.84 & 51.96 & 52.68 & 47.55  & 50.76      & 66.40  \\
                            & +both                     & 47.09 & 53.17 & 53.26 & 50.73  & 56.55      & -  \\
    \midrule
    \multirow{5}{*}{Soft-CE}
                            & baseline                  & 40.49 & -     &  -    & 35.23  & -      & 58.07  \\
                            & +answer mask (sigmoid)    & 39.32 & -     &   -   & 35.21  & -      & -  \\
                            & +answer mask (softplus-G) & 40.30 & -     &   -   & 34.80  & -      & -  \\
                            & +loss re-scaling          & 55.93 & -     &   -   & 52.05  & -      & 69.37  \\
                            & +both                     & 57.02 & -     &   -   & 53.50  & -      & -  \\
    \bottomrule
  \end{tabular}}
\end{table*}

\begin{figure}
  \centering
  \includegraphics[width=0.95\linewidth]{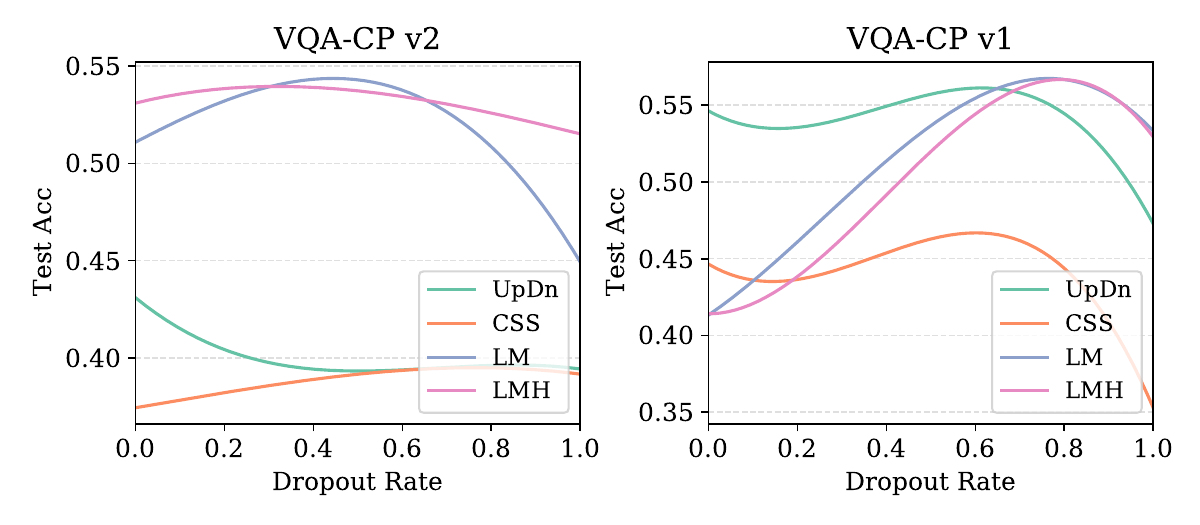}
  \caption{Testing Acc of  four evaluated methods over two datasets with respect to the dropout rate.}\label{fig:dropout}
\end{figure}

\begin{figure}
  \centering
  \includegraphics[width=0.95\linewidth]{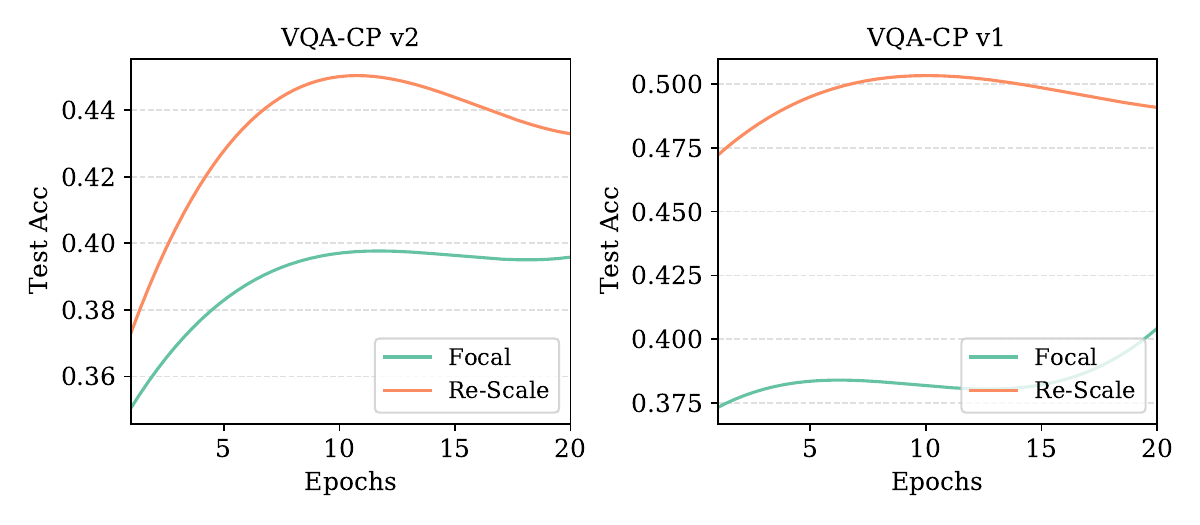}
  \caption{Testing Acc of loss re-weighting methods on the UpDn baseline over two datasets with respect to training epochs.}\label{fig:loss_step}
\end{figure}

\begin{figure*}
  \centering
  \includegraphics[width=0.95\linewidth]{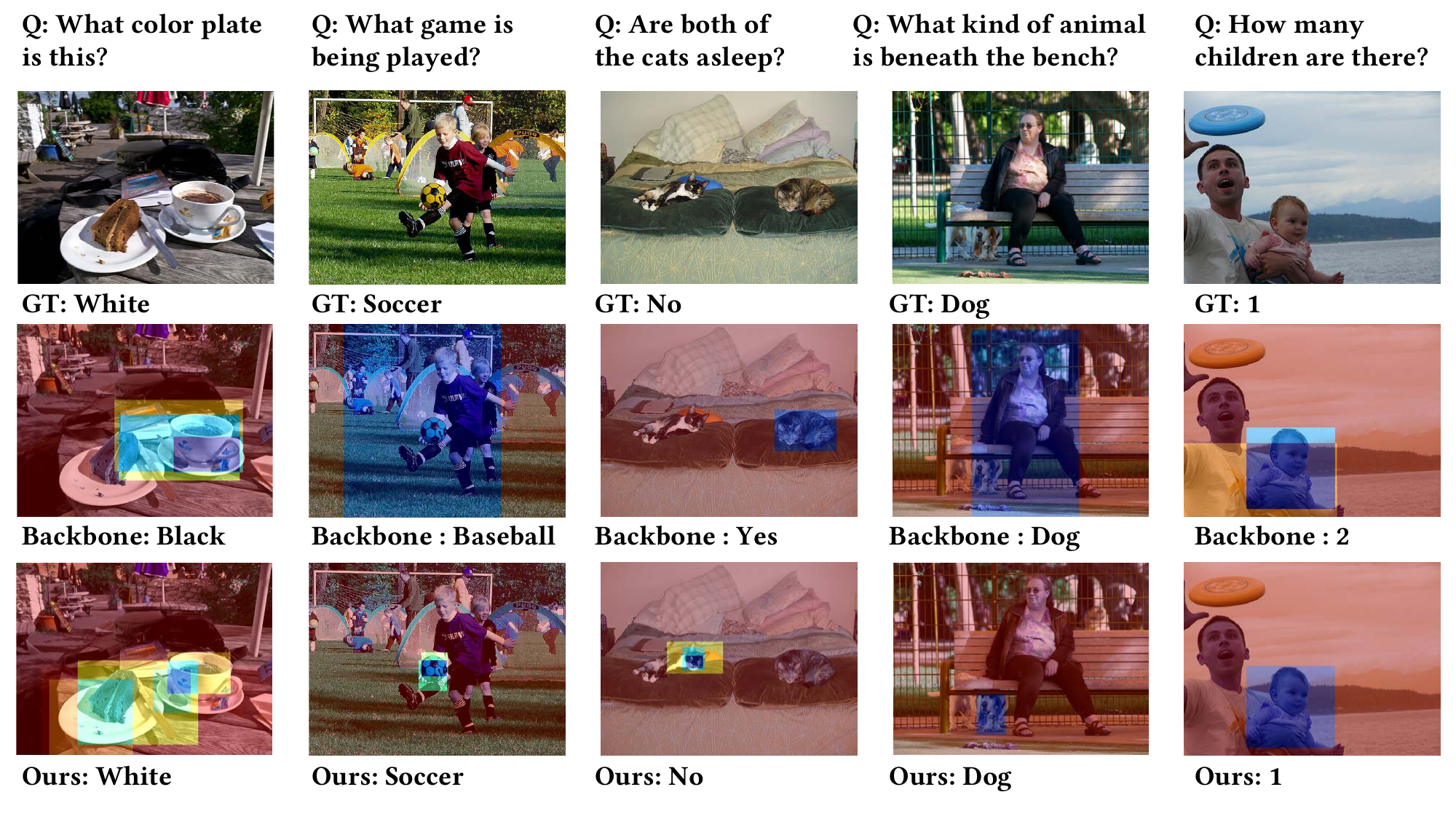}
  \caption{Visualization results from  UpDn and our method on the VQA-CP v2 dataset. Regions with lighter colors denote what the model focuses on when making prediction.}\label{fig:case_study}
\end{figure*}
We conducted detailed experiments to validate the effectiveness of each module and reported the results in Table~\ref{tab:ablation} and Figure~\ref{fig:dropout}. Note that the LM and LMH are particularly developed based on BCE loss function which cannot be adapted with the Soft-CE solution. And we did not apply the answer mask to LXMERT for avoiding undermining its pre-trained linguistic features. We can find that:
\begin{itemize}
  \item The loss re-scaling scheme is effective under a series of settings. When directly applying it to the backbones, the accuracy is enhanced with a large margin. For instance, CSS and LXMERT achieve 8.6\% and 14.62\% absolute gains under the BCE loss, respectively. With the pre-training of the answer mask module, our method can further obtain a higher accuracy.
  \item For the answer mask module, the softplus-G activation function does surpass the sigmoid one. The reason is that the mask tends to be larger than 0.5 with a high probability, making less answers out of the corresponding answer set in the training set.
  \item We further evaluated the dropout rate effect in the answer mask module and illustrated the results in Figure~\ref{fig:dropout}. It can be observed that UpDn prefers a smaller dropout rate, while LM, CSS and LMH fit larger ones.
  \item Surprisingly, our methods can attain a substantially better results over the backbones when using the CE loss function. For example, the final absolute performance improvement of our method over the UpDn backbone is over 16.53\%! It is even better than the recent strong baselines LMH and CSS+LMH. This highlights the effectiveness of the loss re-scaling scheme and the potential of the class-imbalance view on alleviating the language prior problem in VQA.
\end{itemize}
\subsection{Other Loss Re-weighting Method (RQ3)}
To examine whether the loss re-scaling scheme is superior than other commonly used loss re-weighting approaches, we adopted the Focal Loss~\cite{focal, tip-focal} and tested its performance in VQA. Focal Loss is designed to put more focus on hard, mis-classified examples, where the loss weights are auto-learned from the training data, 
\begin{equation}\label{equ:focal}
  FL(p_t) = -\alpha(1 - p_t)^\gamma log(p_t),
\end{equation}
where $\alpha$ and $\gamma$ are hyper-parameters and $p_t$ denotes the prediction for binary classification. We extended this loss function to the VQA multi-class case and plotted the results in Figure~\ref{fig:loss_step}. We can observe that there exists a quite large performance margin between our method and the Focal Loss. One possible reason is that our loss weights is particularly computed under each question type, which provides more explicit guidance than the auto-learning strategy in Focal Loss.
\subsection{Case Study (RQ4)}
We visualized the attention map and predicted answers of UpDn and our method in Figure~\ref{fig:case_study}. Specifically, for the first, second and the last instances, the backbone all provides the incorrect answers due to the language priors. The reason is that  answers \emph{black}, \emph{baseball} and \emph{2} are the dominated answers for question types \emph{what color}, \emph{what} and \emph{how many}, respectively. For the last instance, the attention map from the baseline focuses on both the \emph{man} and the \emph{child}, which misleads the model to predict that there are \emph{2} children. In contrast, our method attends solely on the \emph{child} object and yields the correct answer. For the third instance, the baseline only looks an asleep \emph{cat} and concludes that both cats are asleep, while our method notices the awake \emph{cat}. And for the fourth instance, though the baseline also predicts the right answer, nevertheless, its attention is focused on both the \emph{women} and the \emph{dog}. Instead, our method puts more attention on the animal beneath the bench - \emph{dog}.

\section{Conclusion and future work}\label{conclusion}
There are few studies on analyzing the inherent cause of the language prior problem in VQA. In this work, we propose to fill this gap via interpreting it from a class-imbalance view. To begin with, we present two assumptions and empirically validate their viability. We then develop a simple yet effective loss re-scaling approach to attach distinct weights  to each answer according to the given question type. Extensive experiments on three publicly available datasets validate the effectiveness of the proposed method. 

In the future, we would like to exploit other typical methods overcoming the class-balance issue, i.e., data re-balancing or transfer learning, for alleviating the language prior problem. Moreover, the class-imbalance view  can be generalized to other vision-and-language tasks which are also hindered by the language bias problem, e.g., image captioning.  

\appendices
\section{Losses and Partial Derivatives}\label{appendix}
Based on the re-scaling weight, we further derive the losses and its partial derivatives over the prediction for both the Soft-CE and Sigm-BCE below. With this, the gradient of other learnable parameters can be easily computed.

1) For Sigm-BCE, the loss function becomes,
\begin{equation}\label{equ:loss_bce_new}
  \mathcal{L}_{bce}^\mu = -  \sum_{i = 1}^{|A|} \mu_i( a_i \log \sigma(p_i) + (1 - a_i) \log (1 - \sigma(p_i))),
\end{equation}
where the partial derivative can be obtained by,
\begin{equation}\label{equ:grad_bce_new}
  \begin{aligned}
  \frac{\partial \mathcal{L}_{bce}^\mu}{\partial p_i}   &= - \mu_i (a_i \frac{1}{\sigma(p_i)} (\sigma(p_i)(1 - \sigma(p_i))) \\
                                                        & + (1 - a_i) \frac{1}{1 - \sigma(p_i)} (-\sigma(p_i)(1- \sigma(p_i))))\\
                                                        &= \mu_i((1 - a_i) \sigma(p_i) -  a_i (1 - \sigma(p_i))).
  \end{aligned}
\end{equation}
2) For the Soft-CE, the loss function is given as,
\begin{equation}\label{equ:loss_ce_new}
  \begin{aligned}
  \mathcal{L}_{ce}^\mu &= - \sum_{i = 1}^{|A|} \mu_i a_i \log \frac{\exp p_i}{\sum_{j=1}^{{A}} \exp p_j}.
  \end{aligned}
\end{equation}
To compute the partial derivative $p_i$ of from $\mathcal{L}_{ce}$, we firstly provide the partial derivative of the Softmax function,
\begin{equation}\label{equ:derivative_softmax}
  \frac{\partial \hat{p}_i}{\partial p_i} =
  \begin{cases}
    \hat{p}_i (1- \hat{p}_j), & \mbox{if } i == j \\
    - \hat{p}_i \hat{p}_j, & \mbox{otherwise}.
  \end{cases}
\end{equation}
And then the partial derivative is given by,
\begin{align*}
\frac{\partial \mathcal{L}_{ce}^\mu}{\partial p_i}   &= - \sum_{i = 1}^{|A|} \mu_i a_i \frac{\partial \log \hat{p}_i}{\partial p_i} \\
                                                    &= - \sum_{i = 1}^{|A|} \mu_i a_i \frac{1}{\hat{p}_i} \frac{\partial \hat{p}_i}{\partial p_i} \\
                                                    \stepcounter{equation}\tag{\stepcounter{equation}\theequation} 
                                                    &= \mu_i a_i (\hat{p}_i - 1) + \sum_{k \neq i}^{|A|} \mu_k a_k \hat{p}_i \\
                                                    &= (\mu_i a_i + \sum_{k \neq i}^{|A|} \mu_k a_k) \hat{p}_i - \mu_i a_i \\
                                                    &= \sum_{k=1}^{|A|} \mu_k a_k \hat{p}_i - \mu_i a_i, 
\end{align*}
where line 3 is derived from Equation~\ref{equ:derivative_softmax}. In this way, the final loss becomes,
\begin{equation}\label{equ:loss_overall}
  \mathcal{L} = \mathcal{L}_{cls} + \mathcal{L}_{mask},
\end{equation}
where the $\mathcal{L}_{cls}$ can be either $\mathcal{L}_{ce}^\mu$ or $\mathcal{L}_{bce}^\mu$.

%
\IEEEpeerreviewmaketitle



\ifCLASSOPTIONcaptionsoff
  \newpage
\fi



%

\bibliographystyle{reference/IEEEtran}
\bibliography{reference/class-imbalance-VQA}

\begin{IEEEbiography}[{\includegraphics[width=1in,height=1.25in,clip,keepaspectratio]{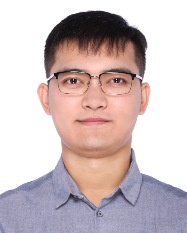}}]{Yangyang Guo}
is currently pursuing the Ph.D. degree with the School of Computer Science and Technology, Shandong University. He will work as a research fellow with the National University of Singapore. He has published several papers in top conferences and journals such as ACM MM, IEEE TKDE. He has served as a Regular Reviewer for journals, including IEEE TMM, IEEE TKDE, ACM ToMM.

\end{IEEEbiography}
\begin{IEEEbiography}[{\includegraphics[width=1in,height=1.25in,clip,keepaspectratio]{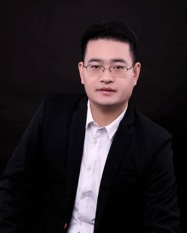}}]{Liqiang Nie} is currently a professor with Shandong University. He received his B.Eng. and Ph.D. degree from Xi'an Jiaotong University and National University of Singapore (NUS), respectively. After PhD, Dr. Nie continued his research in NUS as a research follow for three years. His research interests lie primarily in multimedia computing and information retrieval. Dr. Nie has co-/authored more than 200 papers and 4 books, received more than 12,000 Google Scholar citations. He is an AE of IEEE TKDE, IEEE TMM, ACM ToMM, and Information Science. Meanwhile, he is an area chair of ACM MM 2018-2021. He has received many awards, like ACM MM and SIGIR best paper honorable mention in 2019, SIGMM rising star in 2020, TR35 China 2020, DAMO Academy Young Fellow in 2020, and SIGIR best student paper in 2021.

\end{IEEEbiography}
\begin{IEEEbiography}[{\includegraphics[width=1in,height=1.25in,clip,keepaspectratio]{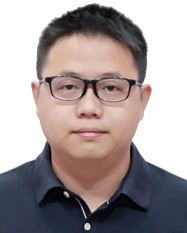}}]{Zhiyong Cheng}
received the M.S. degree from Xi’an Jiaotong University in 2010 the Ph.D. degree in computer science from Singapore Management University in 2016. He was a Research Fellow with the National University of Singapore. He is currently a Professor with the Shandong Artificial Intelligence Institute, Shandong Academy of Sciences, Qilu University of Technology. His research interests mainly focus on large-scale multimedia content analysis and retrieval. He has served as a Regular Reviewer for journals, including TIP, TKDE and TMM.
\end{IEEEbiography}
\begin{IEEEbiography}[{\includegraphics[width=1in,height=1.25in,clip,keepaspectratio]{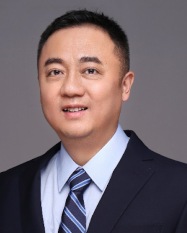}}]{Qi Tian}
received the BE degree in electronic engineering from Tsinghua University and the Ph.D. degree in ECE from the University of Illinois at Urbana-Champaign (UIUC), USA. He is a chief scientist of computer vision in the Huawei Noah’s Ark Laboratory, a full professor with the Department of Computer Science, The University of Texas at San Antonio (UTSA), a Changjiang Chaired professor of the Ministry of Education, an Overseas Outstanding Youth, an Overseas Expert by the Chinese Academy of Sciences, and an IEEE fellow. He was also a visiting chaired professor with the Center for Neural and Cognitive Computation, Tsinghua University, and a lead researcher with the Media Computing Group at Microsoft Research Asia (MSRA).
\end{IEEEbiography}

\begin{IEEEbiography}[{\includegraphics[width=1in,height=1.25in,clip,keepaspectratio]{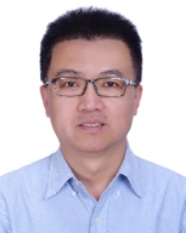}}]{Min Zhang} is currently a professor with Harbin Institute of Technology (Shenzhen), Shenzhen, China. He received his Bachelor degree and Ph.D. degree in computer science from Harbin Institute of Technology, Harbin, in 1991 and 1997, respectively. His current research interests include natural language processing, multi-model, artificial intelligence and cryptology.
\end{IEEEbiography}

\end{document}